\documentclass[10pt, a4paper]{article}
\usepackage{lrec2022} % this is the new LREC2022 Style
\usepackage{multibib}
\newcites{languageresource}{Language Resources}
\usepackage{graphicx}
\usepackage{tabularx}
\usepackage{soul}
\usepackage{titlesec}

\titleformat{\section}{\normalfont\large\bfseries\center}{\thesection.}{1em}{}
\titleformat{\subsection}{\normalfont\SmallTitleFont\bfseries\raggedright}{\thesubsection.}{1em}{}
\titleformat{\subsubsection}{\normalfont\normalsize\bfseries\raggedright}{\thesubsubsection.}{1em}{}
\renewcommand\thesection{\arabic{section}}
\renewcommand\thesubsection{\thesection.\arabic{subsection}}
\renewcommand\thesubsubsection{\thesubsection.\arabic{subsubsection}}
\usepackage{epstopdf}
\usepackage[utf8]{inputenc}

\usepackage{hyperref}
\usepackage{xstring}

\usepackage{color}

\usepackage{subcaption}
\usepackage{csquotes}

\title{BAN-Cap: A Multi-Purpose English-Bangla Image Descriptions Dataset}

\name{Mohammad Faiyaz Khan\textsuperscript{$\top$}$^{\ast}$,
      S.M. Sadiq-Ur-Rahman Shifath\textsuperscript{$\top$}$^{\ast}$,
      Md Saiful Islam$^{\ast}{\dagger}$
      } 

\address{$^{\ast}$Shahjalal University of Science and Technology, Sylhet, Bangladesh, \\
\{mfaiyazkhan, sm01\}@student.sust.edu \\
        $^{\dagger}$University of Alberta, Edmonton, Canada \\
          mdsaifu1@ualberta.ca}

\abstract{
As computers have become efficient at understanding visual information and transforming it into a written representation, research interest in tasks like automatic image captioning has seen a significant leap over the last few years. While most of the research attention is given to the English language in a monolingual setting, resource-constrained languages like Bangla remain out of focus, predominantly due to a lack of standard datasets. Addressing this issue, we present a new dataset \textit{BAN-Cap} following the widely used Flickr8k dataset, where we collect Bangla captions of the images provided by qualified annotators. Our dataset represents a wider variety of image caption styles annotated by trained people from different backgrounds. We present a quantitative and qualitative analysis of the dataset and the baseline evaluation of the recent models in Bangla image captioning. We investigate the effect of text augmentation and demonstrate that an adaptive attention-based model combined with text augmentation using Contextualized Word Replacement (CWR) outperforms all state-of-the-art models for Bangla image captioning. We also present this dataset's multipurpose nature, especially on machine translation for Bangla-English and English-Bangla. This dataset and all the models will be useful for further research.
 \\ \newline  }

\begin{document}

\maketitleabstract

\begingroup\renewcommand\thefootnote{$\top$}
\footnotetext{These authors contributed equally to this work}
\endgroup

\section{Introduction}

Image captioning, a variety of multimodal machine learning, is a research area that integrates and models data from different modalities. It generates humanoid descriptions of images by identifying and analysing their contents. It is more involved than other computer vision or natural language processing tasks because it requires object recognition, the inference of their relationships, and the generation of a meaningful and relevant interpretation using a sequence of words. It leverages the heterogeneity and the correlation of data of different modalities to achieve some original goals of artificial intelligence. It has a wide range of applications. For example, an image captioning system can be used in human-computer interaction, develop a hearing-aid system for visually impaired people, perform concept-based image indexing for information retrieval, automate self-driving cars, and many more.

Because of the availability of large-scale image-sentence pair datasets like Flickr8k \cite{flickr8k}, Flickr30k \cite{flickr30k}, and MS COCO \cite{mscoco}, research interest in image captioning and similar domains has seen an enormous rise in the last decade. The recent advancements in deep learning have made image captioning a sought-after topic for the research community. Deep learning-based models like \newcite{show_and_tell} introduced significant improvement over the traditional machine learning-based models by following the encoder-decoder architecture, which leverages the widespread sequence generation capability of the Recurrent Neural Network (RNN) conditioned by the image. Later attention-based models like \newcite{show-attend-tell} contained a mechanism that attempted to filter only the necessary features from the image while generating captions. Also, multilingual image description datasets are widely available for other languages. The IAPR TC-12 dataset \cite{Grubinger06theiapr} has a collection of 20,000 images and a text caption corresponding to each image in up to three different languages (English, German and Spanish). \newcite{funaki-nakayama-2015-image} curated Japanese translations for the English sentences of the Pascal dataset, which contains 1000 images. \newcite{elliott-etal-2016-multi30k} proposed an image description dataset with English-German sentence pairs. It is an extension of the Flickr30k \cite{flickr30k} dataset that contains 31,014 German translations over a subset of English descriptions and 155,070 German descriptions, which are crowd-sourced independently of the original English descriptions.

\begin{figure}
	\includegraphics[width=.98\linewidth]{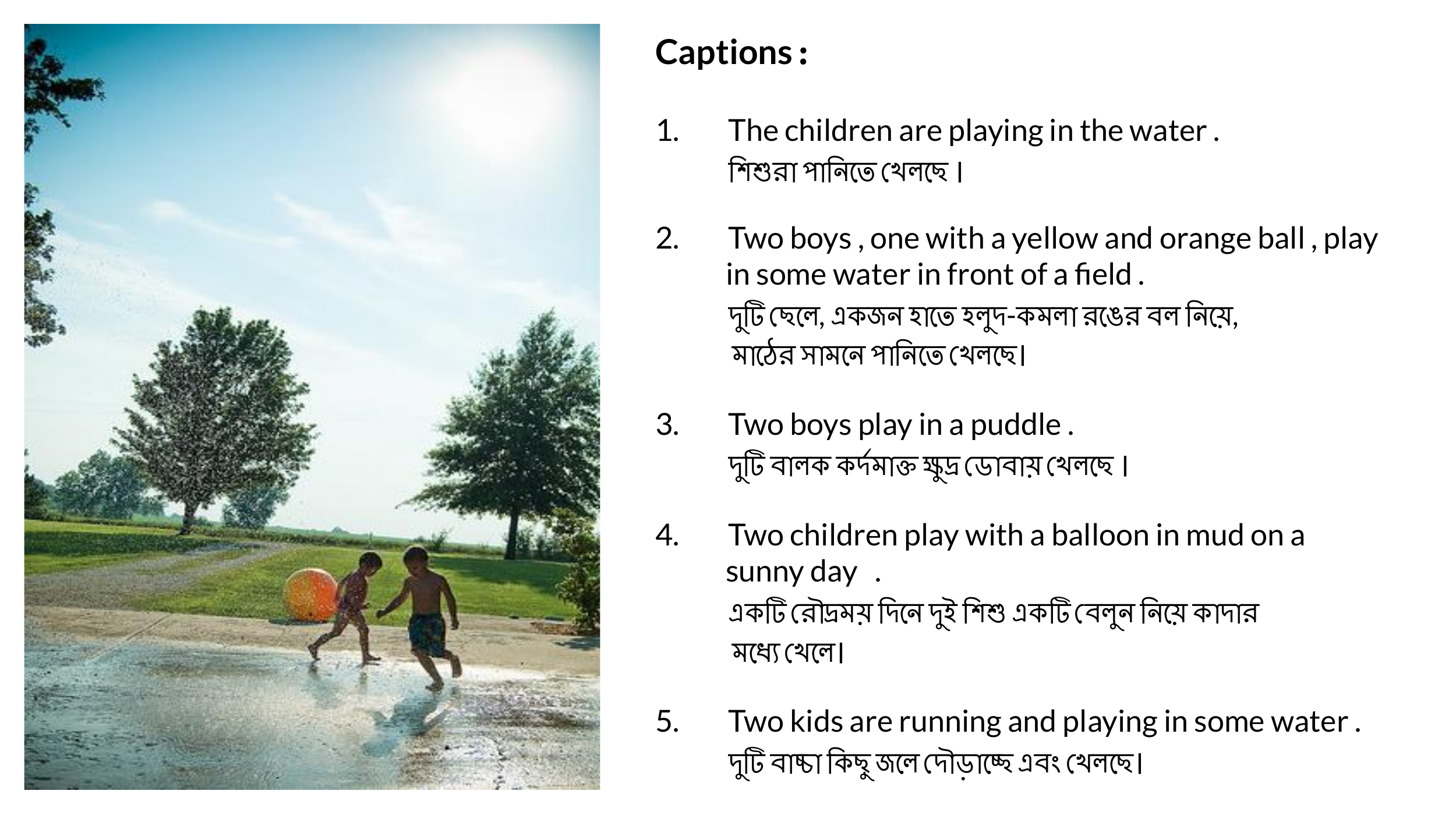}
	\caption{A sample of the dataset containing image and English-Bangla caption pairs}
	\label{fig:ds_example}
\end{figure}

Despite being the fifth most spoken language in terms of the number of speakers\footnote{\url{https://en.wikipedia.org/wiki/List_of_languages_by_number_of_native_speakers}}, Bangla still lacks the availability of a sizable and high-quality image-sentence dataset. So, research in the related fields in the Bangla language is still in its infancy. Some attempts have been made to create image captioning datasets in Bangla. \newcite{chittron} proposed an encoder-decoder model along with a Bangla image captioning dataset \citelanguageresource{banglalekhaimagecaptions}. The dataset contains 9,154 images and two captions per image. It lacks adequate captions per image and contains a number of samples with generic captions which do not provide necessary details. \newcite{Deb2019OboyobAS} presented a comparative analysis of the CNN-LSTM based methods on a sub-sampled, machine-translated version of the Flickr8k \cite{flickr8k} dataset. The resulted dataset lacks variety, quality, and usability. Recently, \newcite{bornon} proposed a transformer-based model along with 4000 images and five human-annotated captions per image. Nevertheless, it has relatively shorter descriptions of an image which result in less detail and lacks usability in domains like multimodal machine translation.

We introduce a sizable dataset of images paired with sentences in English and Bangla to mitigate these problems. It is an extension of the Flicrkr8k \cite{flickr8k} dataset with 8091 images and 40455 English-Bangla caption pairs. The annotations are provided by native Bangla speakers who have expertise in the English and the Bangla language. The dataset is post-processed and evaluated by an expert team for quality maintenance. Figure \ref{fig:ds_example} shows a sample of this dataset with an image and its corresponding English-Bangla caption pairs. Additionally, we demonstrate this datasets' usability and efficacy by training and evaluating multiple deep learning-based methods for image description generation and machine translation. Experimental results show our dataset's variety and diversity and validate its multipurpose nature. To our knowledge, this is the first human-annotated image description dataset containing English-Bangla caption pairs. The dataset and code are available at github\footnote{\url{https://github.com/FaiyazKhan11/BAN-Cap}}.

Our contributions are the following:
\begin{itemize}
  \item We present BAN-Cap, an extension of the Flickr8k \cite{flickr8k} image descriptions dataset, by accumulating Bangla captions with competent annotators under various quality control measures to ensure its quality and usability.
  \item We perform a statistical analysis of the data and present a qualitative and quantitative comparison between our human-annotated dataset and a machine-translated dataset curated using Google Translate.
  \item We set up the baseline of this dataset by training and evaluating all the recent notable models of Bangla image captioning. Besides, we present a baseline of neural machine translation to demonstrate this dataset's multipurpose nature.
  \item We present an adaptive attention based model with contextualized word replacement that outperforms current state-of-the-art models in Bangla image captioning. Additionally, we experiment with other text augmentation techniques as a possible direction of improvement the overall models' performances in Bangla image captioning.
  \item We compare the model's prediction on unseen data while trained on our dataset with the models trained on other existing Bangla caption datasets. We present qualitative human evaluation scores of the predictions that show the model trained on our dataset generates quality captions and has better generalization capability.  
\end{itemize}

\begin{table*}[]
\begin{center}
\scalebox{.7}{
    \begin{tabular}{|l|l|l|l|l|l|}
    \hline
    Dataset & \#Sentences & \#Total Tokens & \#Unique Tokens & Sentence Length Mean & Sentence Length Variance \\ \hline
    Flickr8k (English)             & 40455       & 437421         & 8440            & 10.81       & 14.51      \\ \hline
    BAN-Cap (Bangla)              & 40455       & 344574         & 15846          & 8.51         & 10.99     \\ \hline
    BanglaLekhaImageCaptions \citelanguageresource{banglalekhaimagecaptions}    & 18308       & 155249         & 5720          & 8.47         & 20.13     \\ \hline
    Bornon \cite{bornon}                     & 20500       & 110566         & 6228          & 5.34         & 4.38     \\ \hline
    \end{tabular}
}
\caption{Statistics of the textual data of BAN-Cap along with existing Bangla image captioning data}
\label{tab:corpus_statistics}
\end{center}
\end{table*}

\section{The BAN-Cap Dataset}

\subsection{Data Collection}

The Flickr8k dataset contains images collected from a community-based online photo hosting website \cite{flickr8k}. We used the Flickr8k data as it contains evenly distributed images from various domains. Each image has five descriptions in English, which are collected through crowd-sourcing platforms. The BAN-Cap dataset contains Bangla captions of the Flickr8k images provided by human annotators. It has 8091 images and 40,455 English-Bangla description pairs.

\subsubsection{Setup}
Our goal was to minimize various human biases in the annotations throughout the data collection process. We adopted the following procedures:
\begin{itemize}
    \item We divided the annotators into two groups. The first group consisted of twenty native Bangla speakers who studied in the linguistics department at various public universities in Bangladesh. The second group consisted of graduate students with expertise in the Bangla and the English language.
    \item The first group performed the annotation task. The second group provided an overall guideline for the first group and performed error correction and quality evaluation of the annotations.
    \item The gender ratio of males and females in the two groups was 3:2.
    \item The ages of the annotators and the expert group members ranged from 18 to 30.
    \item We ensured that the annotators represented different demographic regions from all over the country.
\end{itemize}
All the above measures were taken to build a group of annotators with as much variety as possible. The members were paid a standard amount depending on their types of work.

% Bangla Sentence length Mean:  8.866172290198987
% Bangla Sentence length Variance:  10.995685390298544

% english Sentence length Mean:  11.717315535780497
% english Sentence length Variance:  14.512513483640326

% Chittron Sentence length Mean:  9.587502731046538
% Chittron Sentence length Variance:  20.132671689783333
% Chittron unique bangla tokens:  5175
% Chittron total bangla tokens:  150121

% Bornon Sentence length Mean:  6.486325162679192
% Bornon Sentence length Variance:  4.382812793960702
% Bornon unique bangla tokens:  5825
% Bornon bangla tokens:  110566

\subsubsection{Human Annotation}
We developed a website for collecting the annotations. The annotators were required to log in using their names and registration numbers. The annotation page contained an image and an English caption. The annotators were asked to provide a Bangla caption primarily based on their understanding of the image and take help from the provided English caption if necessary. The guideline provided to the annotators by the expert group contained instructions like describing the images following the natural flow and native Bangla sentence structure, avoiding transliterated Bangla words as much as possible, using proper punctuation. An annotator provided only one caption for each image which ensures the variety and vibrancy of the data.

\subsubsection{Post Processing}
During the data collection process, the expert group repeatedly assessed the quality of the captions by manually checking a subset of the descriptions randomly. After the annotation phase, they performed manual error correction on the data.

\begin{figure}[ht]
\begin{subfigure}{.5\textwidth}
  \centering
  % include first image
  \includegraphics[width=.8\linewidth]{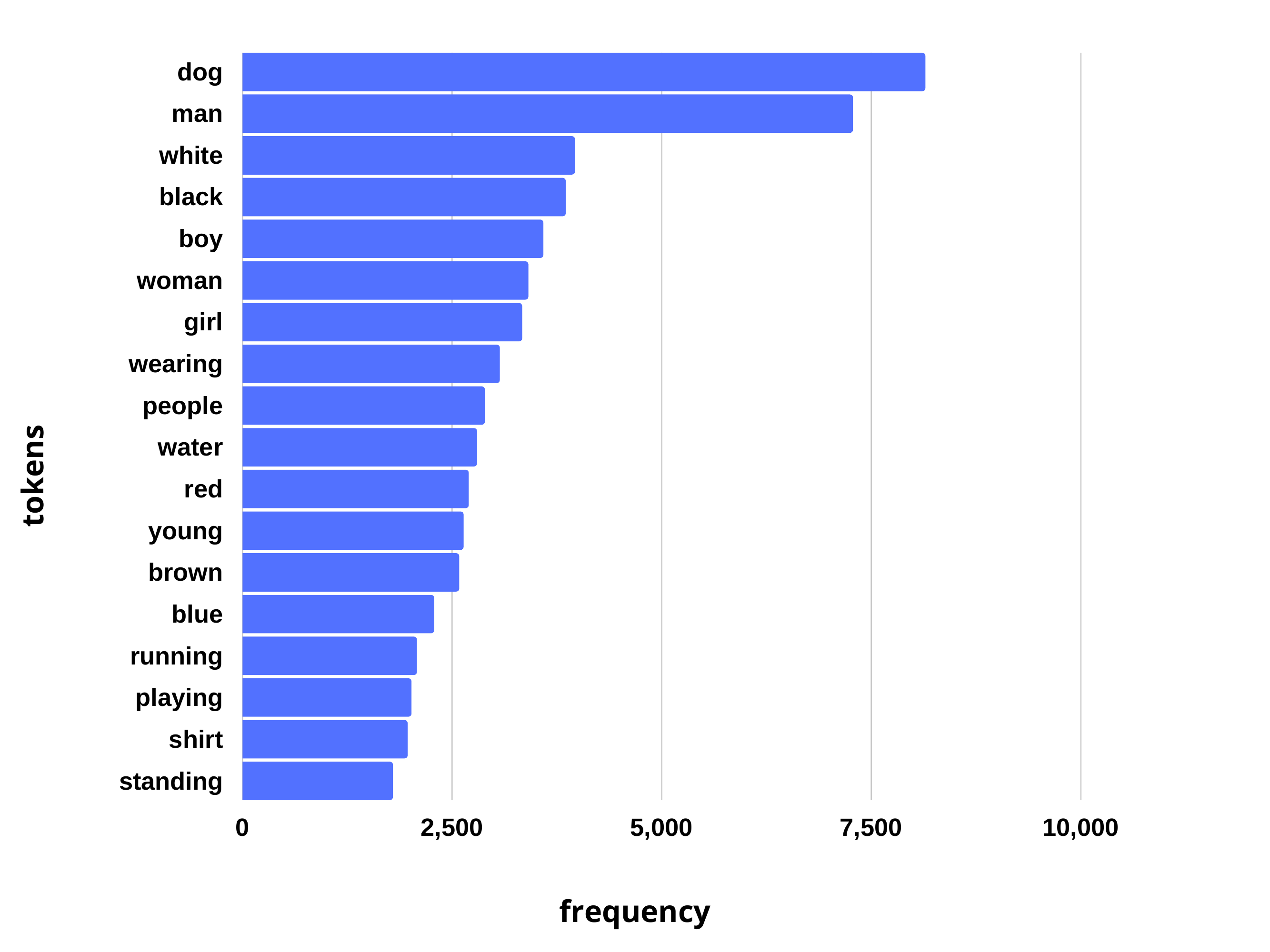}  
  \caption{Most frequent English tokens}
  \label{fig:sub-first}
\end{subfigure}
\begin{subfigure}{.5\textwidth}
  \centering
  % include second image
  \includegraphics[width=.8\linewidth]{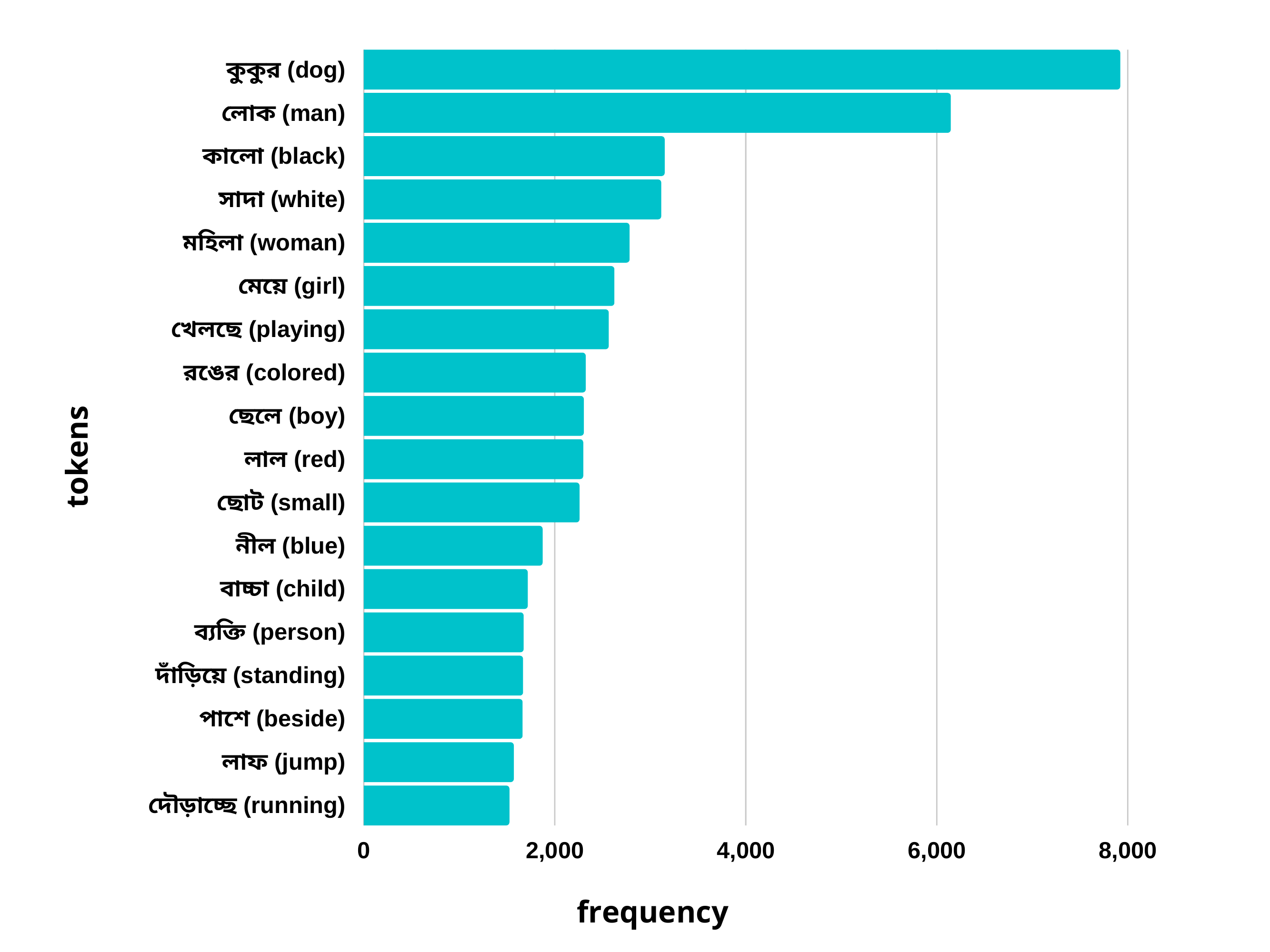}  
  \caption{Most frequent Bangla tokens}
  \label{fig:sub-second}
\end{subfigure}
\caption{Most frequent English and Bangla tokens (Descending order)}
\label{fig:token_frequency}
\end{figure}

\subsection{Statistical Analysis}

Table \ref{tab:corpus_statistics} shows corpus-level statistics and comparison among BAN-Cap and other existing datasets in Bangla. BAN-Cap has higher unique tokens compared to other existing datasets. It has a similar average sentence length compared to the BanglaLekhaImageCaptions \citelanguageresource{banglalekhaimagecaptions} while having more than twice as many captions. Also, the recently proposed human-annotated data, Bornon \cite{bornon}, has a significantly lower average sentence length, which is critical for maintaining the details while describing an image. It is also noticeable that there are some structural variations between Bangla and the English captions. BAN-Cap Bangla descriptions have about 87\% more unique tokens compared to English. On the other hand, the total number of tokens is about 27\% higher in English than in Bangla. Also, an average English description is longer than a Bangla description.

Figure \ref{fig:token_frequency} illustrates the top twenty most frequent tokens in English and Bangla caption datasets. Both languages have several stop words with a high frequency of occurrence. So we discarded the stop-words in this analysis for the sake of comparison. Both the datasets have a similar token frequency.

\begin{figure*}
    \centering
	\includegraphics[width=.8\linewidth]{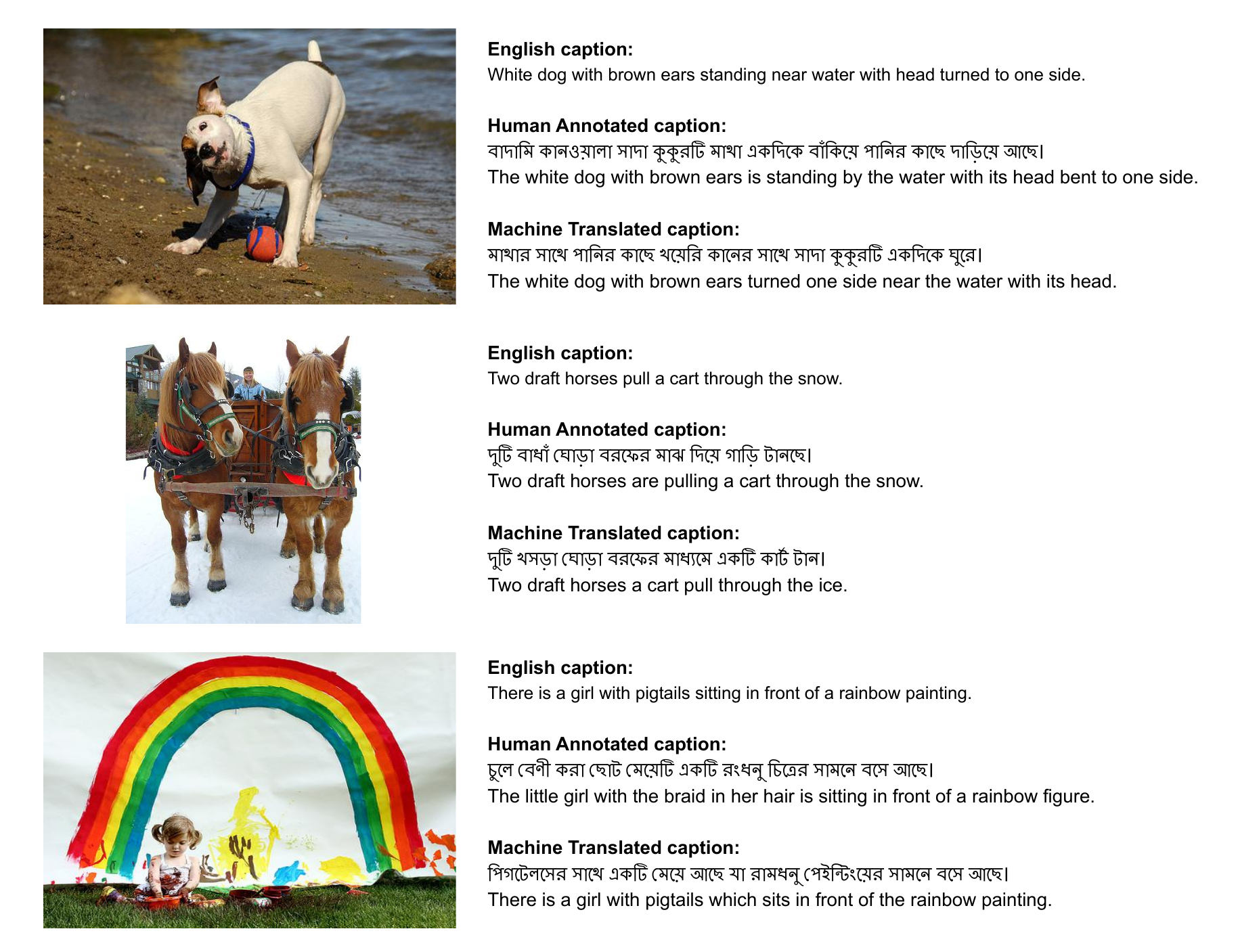} % Figure image
	\caption{Examples where the machine translated captions captured the context of the image but failed to maintain syntactical and structural integrity of Bangla sentence (English translations are provided for the understanding of the non Bangla speakers).} % Figure caption
	\label{fig:annotated_vs_translated_1} % Label for referencing with 
\end{figure*}

\begin{figure*}
    \centering
	\includegraphics[width=.8\linewidth]{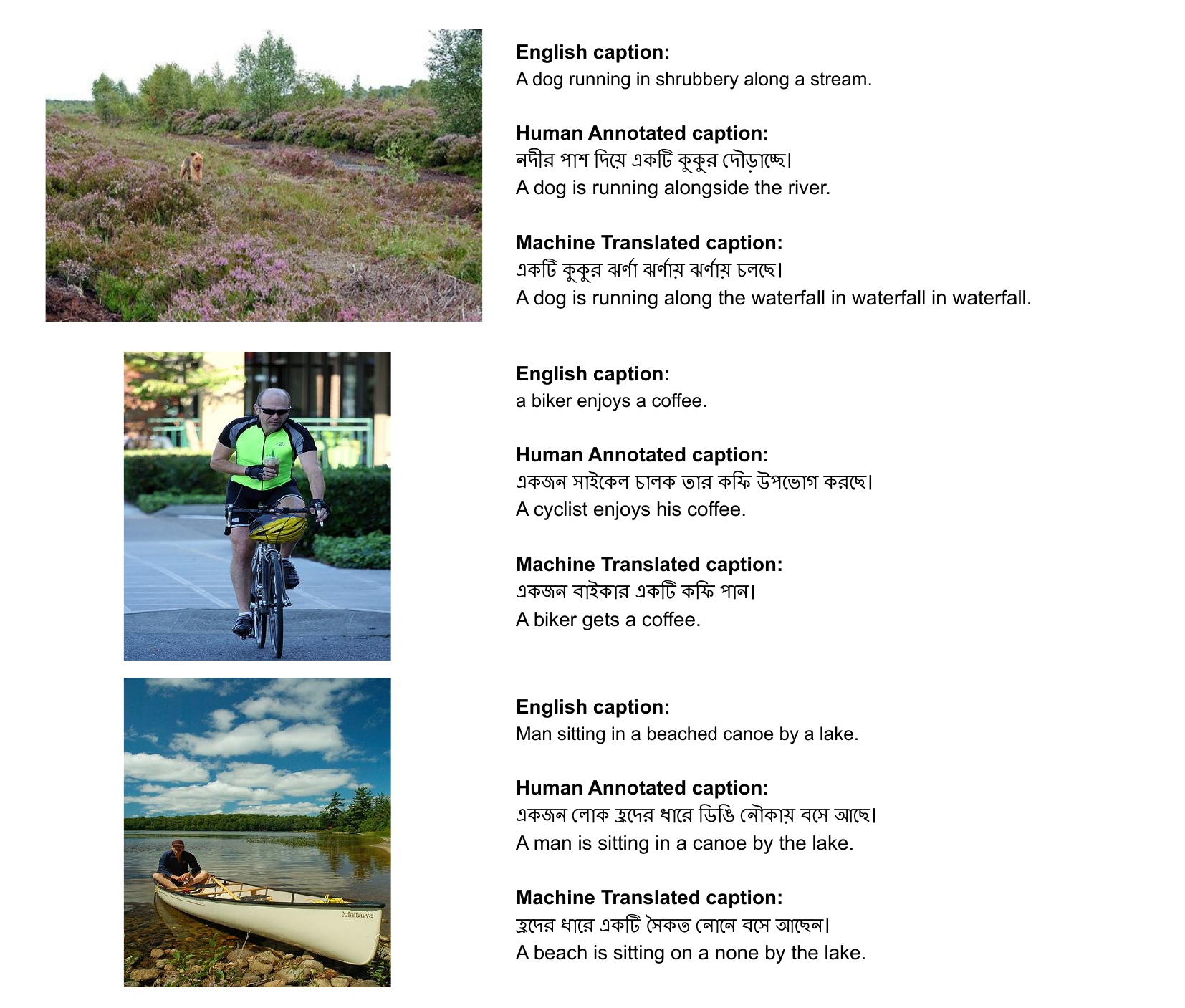} 
	\caption{Examples where the machine translated captions very poorly described the image or failed to generate any meaningful sentence at all (English translations are provided for the understanding of the non Bangla speakers).}
	\label{fig:annotated_vs_translated_2}
\end{figure*}

\subsection{Machine Translated Vs Human Annotated Data}
Automated machine translation tools such as Google Translate have come a long way. However, when it comes to low resource languages like Bangla, they have a considerable amount of lacking. When compared side by side, we observe the following key factors that contribute to curating a human-annotated dataset rather than using a machine-translated one:

\begin{itemize}
    \item The automatic translators are not optimized yet for Bangla. Only a handful of machine-translated captions maintain coherence with the image's content while retaining the structural integrity of a Bangla sentence. It is evident in figure \ref{fig:annotated_vs_translated_1}. We provided English translations for the Bangla captions for the sake of understanding of the non Bangla speaking people.
    
    \item Often the machine-translated Bangla captions contain a tremendous amount of misspelled words, erroneous use of punctuation and incomplete sentences, which do not conclude to a meaningful outcome. These are evident in figure \ref{fig:annotated_vs_translated_2}.
    
    \item The machine-translated captions often fail to capture any cultural essence. From all the examples of figure \ref{fig:annotated_vs_translated_1} and \ref{fig:annotated_vs_translated_2}, it is noticeable that the system has mostly translated the source language into the target language word-by-word. Also, they contain a large amount of transliterated English words in the Bangla captions. So the semantic and the syntactic meaning is lost.
    
    \item The human-annotated captions provide a wide variety compared to the machine-generated captions. In our case, the machine-translated data contains 14606 unique tokens compared to the 15846 tokens in the human-annotated data. However, when we filter out the tokens with at least a frequency of 3, the number of unique tokens in the machine-translated dataset is 4631 compared to 5636 in the human-annotated Bangla dataset.
\end{itemize}

For the above reasons, systems trained with machine-translated data generally output captions which contains artificiality and lack real world usefulness.

% translated unique bangla tokens:  14566, total bangla tokens:  361676

% translated unique bangla tokens:  4631 (347715) (when freq at least 3) human annotated: 5636

\begin{figure}
    \centering
	\includegraphics[width=.9\linewidth]{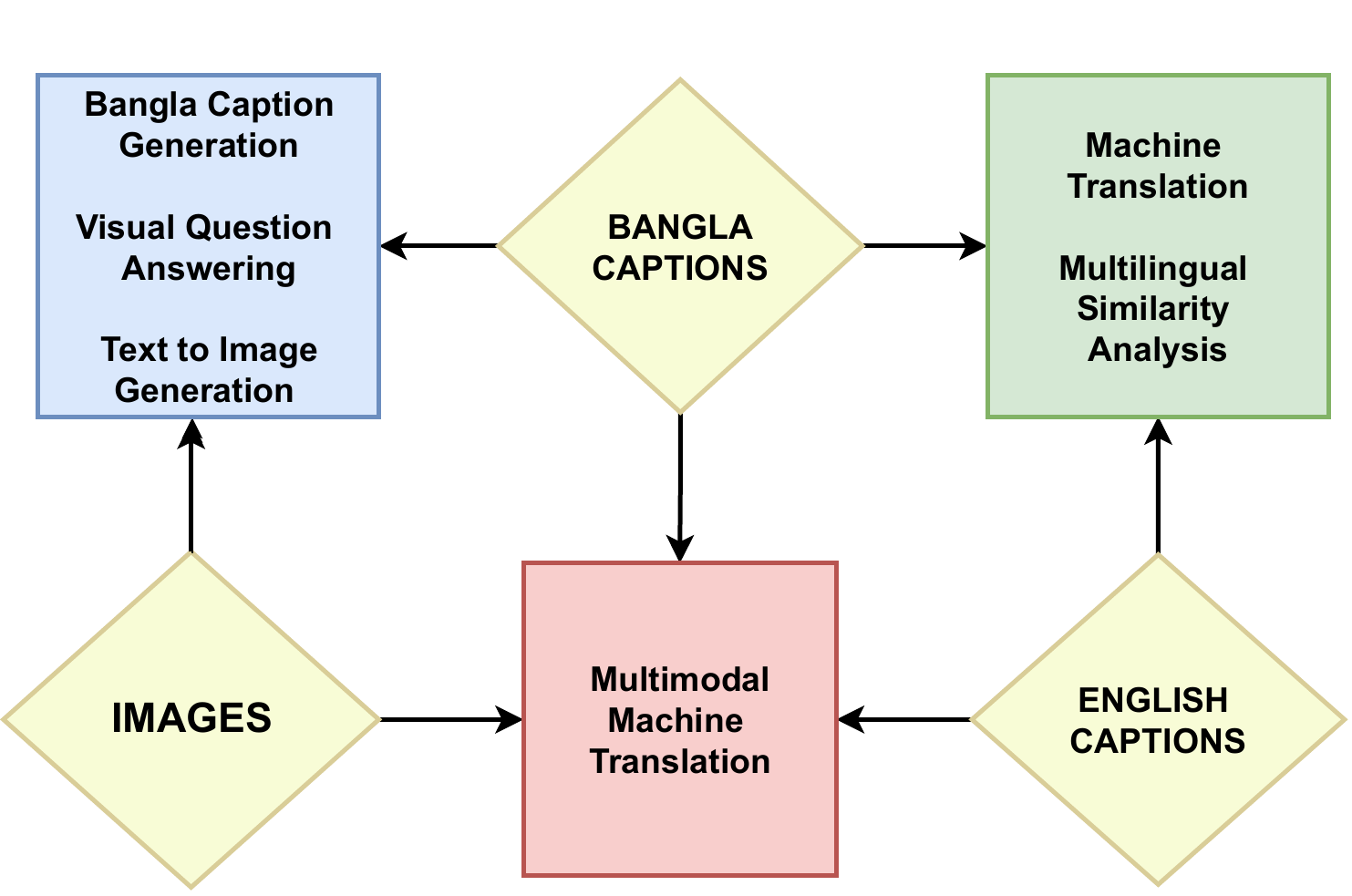} % Figure image
	\caption{A combination of different components of the dataset can be used for different tasks.} % Figure caption
	\label{fig:usecases} % Label for referencing with \ref{bear}
\end{figure}

\subsection{Usability and Multipurpose Nature}

The BAN-Cap dataset can be readily used in various domains along with image captioning. Figure \ref{fig:usecases} illustrates how different components of this dataset can be used for a variety of tasks. Following are some short descriptions of this dataset's usefulness in other domains:

\textbf{Multimodal Machine Translation:} Multimodal Machine translation \cite{mmt_adversarial,mmt_multimodal_trans,mmt_findings,mmt_shared_task,mmt_probing} involves gaining information from multiple modalities. Usually, it is assumed that additional modality features will provide an alternative view of the input data. Unlike machine translation, multimodal machine translation is still a field to explore in the Bangla language. The image-text machine translation can perform better compared to the text-to-text machine translation. It can also be used for cross-modal and cross-lingual information retrieval. 

\textbf{Machine Translation:} Machine translation \cite{nmt_bil,nmt_dev,nmt_global_attn} is already a vastly researched topic in Bangla natural language processing domain. Although there are some well-curated datasets specifically for the English-Bangla machine translation \cite{hasan-etal-2020-low}, the BAN-Cap dataset has its niche with five descriptions containing a context, which eventually carries a diversified view of the same content.

\textbf{Bangla Visual Question Answering:} Given an image and one or more textual questions, a Visual Question Answering (VQA) \cite{vqa_efficacy,vqa_uni_eden,vqa_vlmo} system produces relevant answers by analyzing the questions and the image. It is now a popular research topic across many languages. It can be used to gain information about visual content as well as in tasks like image retrieval. Most of the datasets used in this domain are derived from an image-text dataset like ours. We expect our dataset to play a vital role to kick-start research in the Bangla visual question answering domain.  

\textbf{Text to Image Generation:} Another field of research this dataset can be used is generating an image from text \cite{tti_rifegan2,tti_roberta_attention,tti_zero_shot}. The generated image from a text can serve as a universal language for many applications such as education, language learning, literacy development, summarization of news articles, and data visualization.

\begin{table*}[]
\begin{center}
\scalebox{.8}{
    \begin{tabular}{|l|l|l|l|l|l|l|l|l|}
    \hline
    Model Name                                       & BLEU-1     & BLEU-2   & BLEU-3     & BLEU-4   & CIDEr   & METEOR   & ROUGE$_L$   & SPICE     \\ \hline
    CNN-Merge \cite{improved_bengali}                & 0.565      & 0.355    & 0.221      & 0.131    & 0.178   & 0.281    & 0.290       & 0.042     \\ \hline
    Visual-Attention \cite{bengali_visual_attention} & 0.587      & 0.368    & 0.254      & 0.144    & 0.195   & 0.293    & 0.288       & 0.033     \\ \hline
    Transformer \cite{bornon}                        & 0.623      & 0.396    & 0.251      & 0.152    & 0.198   & 0.300    & 0.290       & 0.038     \\ \hline
    Adaptive-Attention \cite{adaptive_attention}     & 0.702      & 0.466    & 0.307      & 0.194    & 0.297   & 0.297    & 0.344       & 0.055     \\ \hline
    Adaptive-Attention with CWR                      & \textbf{0.738}      & \textbf{0.495}    & \textbf{0.329}      & \textbf{0.208}    & \textbf{0.308}   & \textbf{0.316}    & \textbf{0.368}       & \textbf{0.059}     \\ \hline
    \end{tabular}
}
\caption{Evaluation of different image captioning models on the BAN-Cap dataset.}
\label{tab:image_captioning_main}   
\end{center}
\end{table*}

\begin{table*}[]
\begin{center}
\scalebox{.8}{
    \begin{tabular}{|l|l|l|l|l|l|l|l|l|}
    \hline
    Model Name                                       & BLEU-1     & BLEU-2   & BLEU-3     & BLEU-4   & CIDEr   & METEOR   & ROUGE$_L$   & SPICE     \\ \hline
    CNN-Merge \cite{improved_bengali}                & 0.458      & 0.273    & 0.163      & 0.094    & 0.079   & 0.245    & 0.204       & 0.018     \\ \hline
    Visual-Attention \cite{bengali_visual_attention} & 0.505      & 0.303    & 0.184      & 0.107    & 0.046   & 0.273    & 0.195       & 0.004     \\ \hline
    Transformer \cite{bornon}                        & 0.458      & 0.269    & 0.160      & 0.091    & 0.065   & 0.280    & 0.179       & 0.008     \\ \hline
    Adaptive-Attention \cite{adaptive_attention}                              & 0.515      & 0.310    & 0.188      & 0.109    & 0.055   & 0.267    & 0.198       & 0.005     \\ \hline
    \end{tabular}
}
\caption{Evaluation of different image captioning models on the machine translated Flickr8k Bangla dataset.}
\label{tab:image_captioning_gtranslated}   
\end{center}
\end{table*}

% ============================== 2 caps

\begin{table*}[]
\begin{center}
\scalebox{.8}{
    \begin{tabular}{|l|l|l|l|l|l|l|l|l|}
    \hline
    Model Name                                       & BLEU-1     & BLEU-2   & BLEU-3     & BLEU-4   & CIDEr   & METEOR   & ROUGE$_L$   & SPICE     \\ \hline
    CNN-Merge \cite{improved_bengali}                & 0.468      & 0.279    & 0.167      & 0.096    & 0.059   & 0.256    & 0.198       & 0.013     \\ \hline
    Visual-Attention \cite{bengali_visual_attention} & 0.464      & 0.277    & 0.166      & 0.095    & 0.036   & 0.224    & 0.159       & 0.011     \\ \hline
    Transformer \cite{bornon}                        & 0.466      & 0.277    & 0.166      & 0.095    & 0.035   & 0.255    & 0.142       & 0.013     \\ \hline
    Adaptive-Attention \cite{adaptive_attention}                  & 0.480      & 0.290    & 0.176      & 0.102    & 0.042   & 0.226    & 0.171       & 0.014     \\ \hline
    \end{tabular}
}
\caption{Evaluation of different image captioning models on the BAN-Cap dataset (two  captions per image).}
\label{tab:image_captioning_2_caps}   
\end{center}
\end{table*}

% ====================================

% ============================== 3 caps

\begin{table*}[]
\begin{center}
\scalebox{.8}{
    \begin{tabular}{|l|l|l|l|l|l|l|l|l|}
    \hline
    Model Name                                       & BLEU-1     & BLEU-2   & BLEU-3     & BLEU-4   & CIDEr   & METEOR   & ROUGE$_L$   & SPICE     \\ \hline
    CNN-Merge \cite{improved_bengali}                & 0.483      & 0.281    & 0.166      & 0.094    & 0.010   & 0.143    & 0.192       & 0.002     \\ \hline
    Visual-Attention \cite{bengali_visual_attention}        & 0.484      & 0.291    & 0.174      & 0.100    & 0.034   & 0.237    & 0.167       & 0.018     \\ \hline
    Transformer \cite{bornon}                        & 0.487      & 0.291    & 0.174      & 0.100   & 0.034   & 0.232    & 0.179       & 0.016     \\ \hline
    Adaptive-Attention \cite{adaptive_attention}                             & 0.489      & 0.293    & 0.175      & 0.101    & 0.038   & 0.226    & 0.173       & 0.018     \\ \hline
    \end{tabular}
}
\caption{Evaluation of different image captioning models on the BAN-Cap dataset (three captions per image).}
\label{tab:image_captioning_3_caps}   
\end{center}
\end{table*}

\begin{table*}[]
\begin{center}
\begin{tabular}{|l|l|l|l|l|l|}
\hline
Model             & BLEU-1 & BLEU-2 & BLEU-3 & BLEU-4 & METEOR \\ \hline
Bangla-To-English & 0.610  & 0.375  & 0.229  & 0.134  & 0.132  \\ \hline
English-To-Bangla & 0.656  & 0.419  & 0.264  & 0.158  & 0.306  \\ \hline
\end{tabular}
\caption{Evaluation of different models of machine translation on BAN-Cap dataset.}
\label{tab:machine_translation_evaluation}
\end{center}
\end{table*}
 
\section{Baselining}

\subsection{Existing Models}

\subsubsection{Image Captioning}
To set up the baseline of our dataset, we present the evaluation scores of all the recent state-of-the-art models of Bangla image captioning. We trained the following models on our dataset:

\textbf{CNN-Merge}: \newcite{improved_bengali} proposed this model following the merge architecture of \newcite{tanti-etal-2017-role}. 

\textbf{Visual-Attention}: Proposed by \newcite{bengali_visual_attention}, The visual attention model is very similar to the one introduced in \newcite{show-attend-tell}.

\textbf{Transformer}: This model, proposed by \newcite{bornon}, is also based on the encoder-decoder architecture. It utilizes the multi-head attention of the transformer during decoding.

\subsubsection{Machine Translation}

\textbf{Encoder-Decoder}: This model is the replication of the one proposed in \newcite{nmtsec2sec}

% ====================================

\subsection{A Better Approach}
In a quest to improve the existing baselines, we adopted the highly effective adaptive attention mechanism and trained a similar model existing in the English language \cite{adaptive_attention}. 
\subsubsection{Adaptive - Attention Model}
This model uses \enquote{Sentinel Attention} as an addition to spatial attention. The spatial attention is adaptive in the sense that it is dependent on the current hidden state rather than previous hidden states in \newcite{show-attend-tell}.

The sentinel attention requires a sentinel gate, which determines what kind of information the model will focus on - visual or textual. A context vector is obtained by combining sentinel and spatial visual information. 

\subsubsection{Text Augmentation}

Text augmentation has been a handy technique in many low resource language-based tasks. To investigate if it has an impact on improving the existing models' performances, we experimented with the following three text augmentation techniques in the context of Bangla image captioning:

\textbf{1. Synonymous Word Replacement (SWR)}: Each word of the captions was replaced with a synonymous word using bnltk library\footnote{\url{https://pypi.org/project/bnltk/}} and based on the semantic similarity scores with the human-annotated captions, the top 3 captions were selected.
\cite{ic_text_aug}.

\textbf{2. Back Translation (BT)}:
Each caption was translated to English and then translated back to Bangla using Google Translate\footnote{\url{https://translate.google.com/}}. The back-translated captions were added with the original captions.

\textbf{3. Contextualized Word Replacement (CWR)}:
Each word of the captions was replaced with a contextually similar word predicted by Bangla-BERT\footnote{\url{https://huggingface.co/sagorsarker/bangla-bert-base}} and based on the semantic similarity scores with the human-annotated captions, the top three samples were selected \cite{ic_text_aug}.

\section{Experimentation Details}
\subsection{Data Preprocessing}
For training and evaluation, we split the dataset using the standard train, test and validation split \cite{karpathy_split} for the Flickr8k dataset. We trained on 6000 images, validated on 1000 images, and tested on 1000 images and their corresponding captions. We used only the captions associated with the images in the training, validation, and test sets for machine translation. We applied fundamental preprocessing techniques. We removed punctuations from the captions during tokenization. Each caption begins and ends with a unique starting and ending token. For consistency, captions are either padded or truncated to a fixed size. We set a threshold of five for image captioning and replace all tokens with \enquote{unk} that occur less frequently than five times.
\subsection{Training Process}
% We used publicly available implementations in github\footnote{Image captioning (Soft and adaptive attention): \url{https://github.com/s1879281/Image-Captioning-with-Adaptive-Attention}\\
% Image captioning (Merge and transformer): \url{https://github.com/jelifysh/Image-Captioning}\\
% Machine translation: \url{https://github.com/tejasvi96/Neural-Machine_Translation}} for all the models described above. 
We trained the models on the training data and calculated validation loss on the validation data after each training epoch. The model with the lowest validation loss is saved and later used to predict the unseen test data.
 
\section{Result Analysis}

We evaluated the models' performances using existing evaluation metrics such as BLEU (Bilingual Evaluation Understudy) \cite{bleu}, ROUGE$_L$( Recall-Oriented Understudy for Gisting Evaluation) \cite{rouge}, METEOR (Metric for Evaluation of Translation with Explicit Ordering) \cite{meteor}, CIDEr (Consensus-based Image Description Evaluation) \cite{cider}, and SPICE (Semantic Propositional Image Caption Evaluation) \cite{spice}.  

BLEU is calculated by comparing the reference and predicted sentences' n-gram geometric means. However, because the same sentence can be represented in various ways with the same sense, the scores are not always accurate. The predicted captions were evaluated using 1,2,3, and 4-gram BLEU scores.
ROUGE$_L$ is calculated by comparing the reference and predicted sentences' longest common subsequences. The sentences' longest common subsequence takes sentence-level structure similarity into account and recognizes the longest co-occurrence in n-grams.
CIDEr is calculated using a Term Frequency Inverse Document Frequency (TF-IDF) weighting scheme for the n-gram of each sentence. METEOR is calculated by comparing the actual and predicted sentences word for word and then by calculating the precision and recall harmonic means. SPICE is calculated for sentence pairs based on F-scores on tuples from the scene graphs, semantic representations of the objects, properties, and connections in the captions. CIDER and SPICE are unique metrics for evaluating image captions' syntactic and semantic quality, while BLEU and METEOR are used to assess image captioning and machine translation.

\begin{figure*}
    \centering
	\includegraphics[width=.7\linewidth]{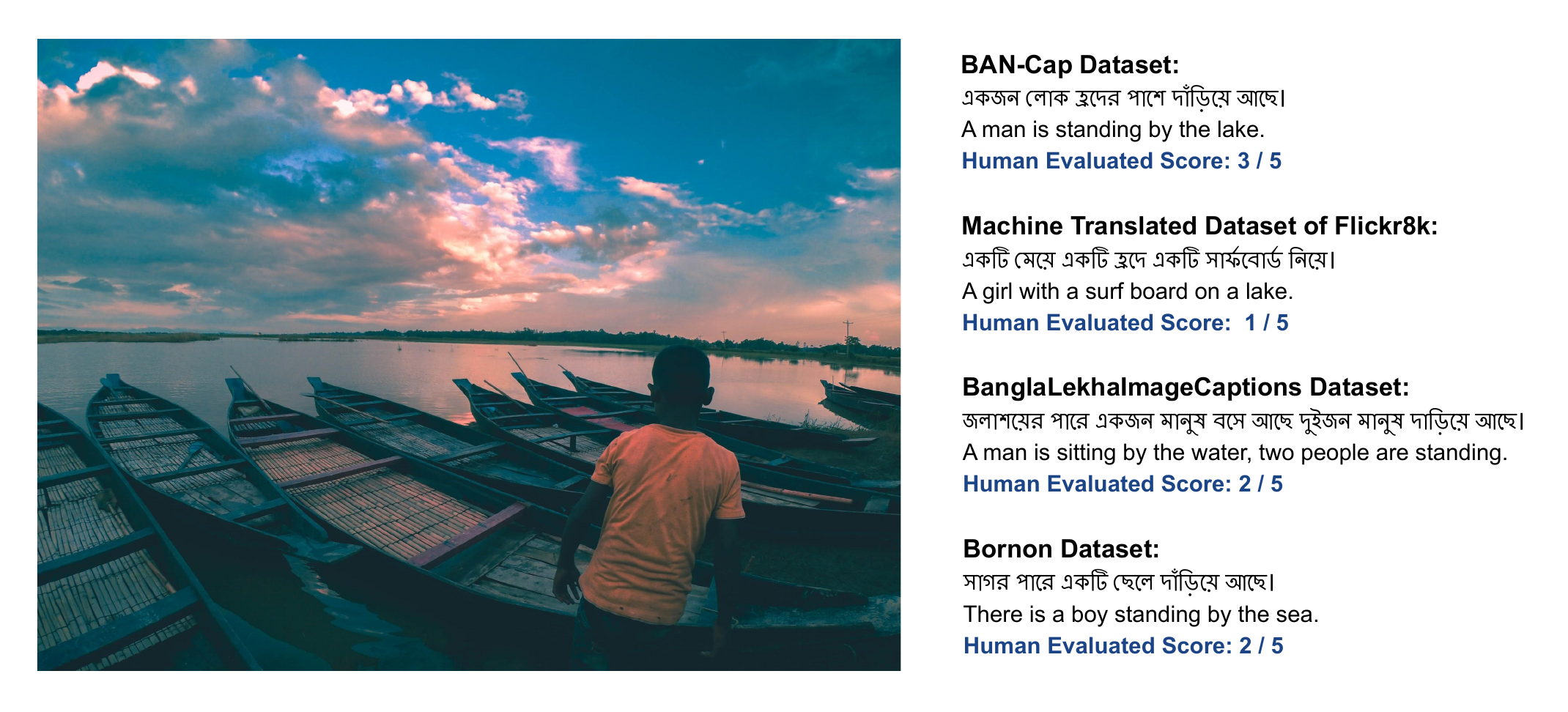} % Figure image
	\includegraphics[width=.7\linewidth]{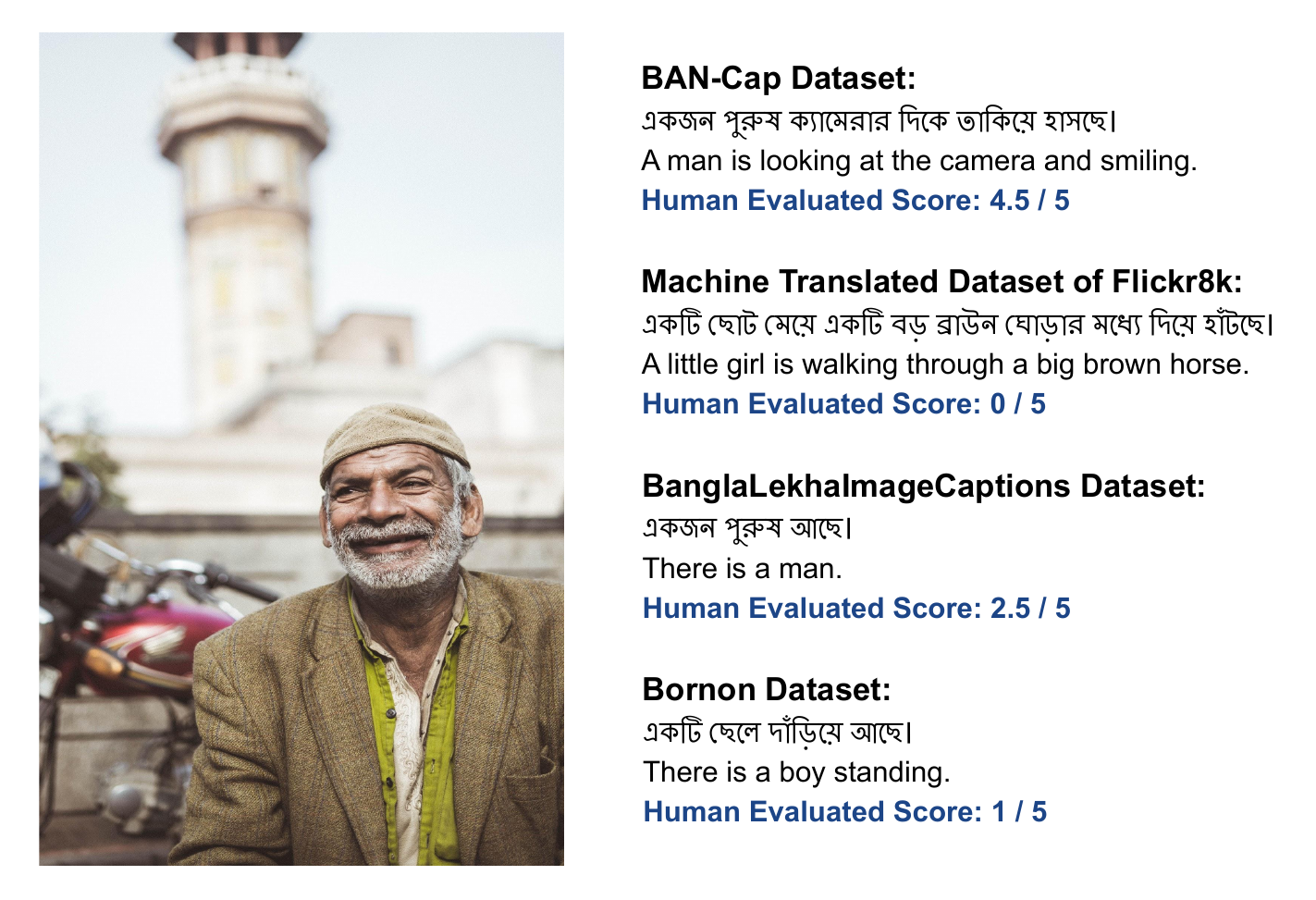} % Figure image
	\caption{Example of the model's prediction on unseen images while trained on different datasets along with corresponding human evaluation scores. (English translations are provided for the understanding of the non native Bangla speakers)} % Figure caption
	\label{fig:human_evaluation} % Label for referencing with 
\end{figure*}

Table \ref{tab:image_captioning_main} contains the evaluation scores of different image captioning models on the test set of the main dataset. In Table \ref{tab:image_captioning_main}, the CNN-Merge \cite{improved_bengali} model achieved lowest scores in all evaluation metrics. The Visual-Attention \cite{bengali_visual_attention} model improves the performance by utilizing the extraction of only important features from an image during a caption prediction. However, despite having a relatively simple architecture, the Transformer \cite{bornon} model outperforms the Visual-Attention and the CNN-Merge models by utilizing multi-head attention and better context awareness ability of the transformer. The adaptive attention-based model outperforms all the other models in most evaluation metrics by applying visual sentinel to guide the model using the attention mechanism more effectively. Finally, we see a performance boost for every model when applying text augmentation. After experimenting with all the combinations of text augmentation techniques previously described, we find that the Adaptive-Attention model with Contextualized Word Replacement gives us the best evaluation scores.     

Table \ref{tab:image_captioning_gtranslated} contains the performance of the image captioning models while trained and evaluated on the machine translated Flickr8k Bangla dataset. All the models show a performance drop across all the metrics. A significant drop can be observed in CIDEr and SPICE, specialized evaluation metrics for image captioning.

We hypothesized that training an image captioning system with more captions per image will yield a better and more robust model that will be able to deliver more varied and detailed captions. To test our hypothesis, we trained the image captioning models with datasets containing two and three captions per image and report the result in Table \ref{tab:image_captioning_2_caps} and \ref{tab:image_captioning_3_caps} respectively. The experimental results indeed validate our hypothesis as we see a gradual improvement in the results from Table \ref{tab:image_captioning_2_caps} to Table \ref{tab:image_captioning_3_caps}. Finally, in Table \ref{tab:image_captioning_main}, we see the highest scores achieved by each model when trained with the dataset containing five captions per image.

Though the metrics mentioned above generally give a numeric estimation of how well a model performs, they often fail to summarise how the predictions appear to a human in real-life use-cases. To get a qualitative idea of how a model predicts unseen images outside the  datasets it has been trained on, we collected some sample images from an online copyright-free source \cite{unsplash_photos}. We trained the best performing Adaptive-Attention model on our dataset, the Google translated Bangla dataset, the BanglaLekhaImageCaptions dataset, and the Bornon dataset. Thus we obtain four different versions of the Adaptive-Attention model. We generated four predictions of each collected image by each of the four versions of the Adaptive-Attention model. Then we asked four experts to assign a score out of five by evaluating the quality of a prediction where a higher score means a better quality caption. The model achieved 3.5/5 on average when trained on our dataset, 2.5/5 on the BanglaLekhaImageCaptions dataset, 2.5/5 on the Bornon, and 1.0/5 on the machine-translated dataset. Figure \ref{fig:human_evaluation} contains samples of the predictions made by the model when trained on different datasets, along with the corresponding human evaluation score.

Table \ref{tab:machine_translation_evaluation} provides the experimental results on the Bangla to English and English to Bangla machine translation task of the encoder-decoder model. Our primary purpose is not to achieve state-of-the-art results but to demonstrate this dataset's multipurpose nature.

\section{Conclusion}

We present BAN-Cap, a multilingual image descriptions dataset containing English-Bangla caption pairs. Expert annotations under intense supervision make it a gold standard dataset. To validate this dataset's multipurpose nature, we test and evaluate it on various models of image captioning and machine translation. We also experiment with text augmentation to add variety to the human-annotated captions. Our future works will include investigating the impact of text augmentations on other existing datasets to validate its generalizability and apply this dataset in different research areas. We expect the proposed dataset will be helpful in the multimodal and multilingual research domain and hope it will be beneficial to the research community for a variety of other purposes that we cannot predict.

\section{Acknowledgements}

We want to thank the annotators and evaluators who helped us in data collection and during the human evaluation process. We would also like to thank Natural Language Processing Group, Department of CSE, SUST for their valuable comments on our work.

% \nocite{*}
\section{Bibliographical References}\label{reference}
%\label{main:ref}

\bibliographystyle{lrec2022-bib}
\bibliography{lrec2022-example}

\section{Language Resource References}
\label{lr:ref}
\bibliographystylelanguageresource{lrec2022-bib}
\bibliographylanguageresource{languageresource}

\section*{Appendix}
\subsection*{Baseline Model Details}
\subsubsection*{Image Captioning}
The general procedure of an image captioning system is generating a sequence of words conditioned by the image and the previously generated words. A convolutional neural network is used to generate image features. The image captioning model tries to find the caption that maximizes the following log probability. 
\begin{equation}
    \log p(S/I) = \sum_{t=0}^{N}\log p(S_t|I,S_o,S_1,S_2.....S_{t-1})
    \label{eqn:word_log_prob}
\end{equation}

Here, $I$ is the image, $S$ is the caption, and $S_t$ is the word in the caption at location $t$. The probability of the word $S_t$ depends on the image $I$ and all the previous words from $S_0$ to $S_{t-1}$. Brief explanations of the models we trained on our dataset for baselining are given in the following sections.

\begin{table*}[]
\begin{center}
\begin{tabular}{|l|l|l|l|l|}
\hline
Model Name         & Batch Size & Learning Rate & Loss Function           & Optimizer \\ \hline
CNN-Merge          & 64         & 0.01          & Cross-Entropy           & Adam      \\ \hline
Soft-Attention     & 32         & 0.0004        & Cross-Entropy           & Adam      \\ \hline
Adaptive-Attention & 32         & 0.0004        & Cross-Entropy           & Adam      \\ \hline
Transformer        & 128        & 0.01          & Cross-Entropy           & Adam      \\ \hline
Encoder-Decoder    & 128        & 0.01          & Negative Log Likelihood & Adam      \\ \hline
\end{tabular}
\caption{Selection of hyperparameters for different models.}
\label{tab:hyperparameters}   
\end{center}
\end{table*}

\subsubsection*{CNN-Merge}
This model is based on the merged architecture proposed in \newcite{tanti-etal-2017-role}
. The image features are encoded using a CNN, and the text features are encoded using another one-dimensional CNN. Here, one dimensional CNN is used instead of  LSTM as it performs better in capturing details of short sentences in Bangla language \cite{improved_bengali}. The image and text feature extraction are independent processes. The two features are merged and passed to a decoder layer for caption generation.

\subsubsection*{Visual-Attention}
 Here, ResNet-101 \cite{resnet} is used to generate feature maps from the image. The relevant location from the feature map is determined, and the location feature vector is passed to the LSTM at each time step. The probability distribution over all the locations is modelled based on the previously generated words. The probability of choosing a location $i$, denoted by $\alpha_{t,i}$ is proportional to the similarity between vector at that location $l_i$ and the LSTM hidden vector $h_t$. The context vector $z_t$ is calculated as 
\begin{equation}
    z_t = \sum_{i=0}^{n}\alpha_{t,i}l_i
    \label{eqn:context_vector}
\end{equation}
Equation \ref{eqn:word_log_prob} changes to the following.
\begin{equation}
    \log p(S/I) = \sum_{t=0}^{N}\log p(S_t|z_t,S_o,S_1,S_2.....S_{t-1})
    \label{eqn:context_log_prob}
\end{equation}
Instead of the whole image, the only relevant context of the image based on the previously generated words is used.

\subsubsection*{Transformer}
A pre-trained Inception-V3 \cite{inceptionv3} extracts the image features. The final classification layer is discarded as only the image vectors are needed. Token and positional embeddings are generated from the captions and passed to a masked multi-head attention layer. This layer's output is passed along with the image features to another multi-head attention layer. The output is routed through a feed-forward layer and then a normalization layer. A softmax layer generates the final output probabilities.

\subsection*{Machine Translation}
\subsubsection*{Encoder-Decoder}
In an encoder-decoder architecture for sequence-to-sequence task \cite{nmtsec2sec}, an encoder reads an input source sentence and generates a vector representation. We used GRU as an encoder.
\begin{equation}
    h_t = f(s_t, h_{t-1})
    \label{eqn:hidden_state}
\end{equation}

\begin{equation}
    v = g(\{h_1, ... , h_{tx}\})
    \label{eqn:vector_representation}
\end{equation}

Here, $s = (s_1, ..., s_{tx})$ is the input sentence, $h_t$ is the hidden state at time $t$, and $v$ is a vector generated from the hidden states. $f$ and $g$ are nonlinear functions. Another GRU is used as a decoder. The decoder predicts the probability of the next word based on the context vector $v$ and all the previously predicted words.  
\begin{equation}
    p(y) = \prod_{t = 1}^{T}p(y_t| \{y_1,..., y_{t-1}\}, v)
    \label{eqn:decoder_prediction}
\end{equation}
where, $y = \{y_1,..., y_t\}$ is the sequence of predicted words.

\subsection*{Training Hyperparameters}
The selection of hyperparameters for training all the models is summarised in Table \ref{tab:hyperparameters}.

\end{document}